# Registration-Free Hybrid Learning Empowers Simple Multimodal Imaging System for High-quality Fusion Detection

Yinghan Guan,[1,†] Haoran Dai,[2,†] Zekuan Yu,[3] Shouyu Wang,[1,4] and Yuanjie Gu [1,3,5]

1 School of Electronics and Information Engineering, Wuxi University, Wuxi 214105, China.
2 Metropolitan College, Department of Computer Science, Boston University, Boston, 02215, USA
3 Academy for Engineering and Technology, Fudan University, Shanghai 200433, China
† Equal contribution
4 shouyu29@cwxu.edu.cn
5 yuanjie_gu@163.com

**Multimodal fusion detection always places high demands on the imaging system and image pre-processing, while either a high-quality pre-registration system or image registration processing is costly. Unfortunately, the existing fusion methods are designed for registered source images, and the fusion of inhomogeneous features, which denotes a pair of features at the same spatial location that expresses different semantic information, cannot achieve satisfactory performance via these methods. As a result, we propose IA-VFDnet, a CNN-Transformer hybrid learning framework with a unified high-quality multimodal feature matching module (AKM) and a fusion module (WDAF), in which AKM and DWDAF work in synergy to perform high-quality infrared-aware visible fusion detection, which can be applied to smoke and wildfire detection. Furthermore, experiments on the M³FD dataset validate the superiority of the proposed method, with IA-VFDnet achieving the best detection performance than other state-of-the-art methods under conventional registered conditions. In addition, the first unregistered multimodal smoke and wildfire detection benchmark is openly available in this letter.**

Multimodal fusion detection implies complementary information, stability, and safety and has long been an important part of disaster detection [1, 2], autonomous driving perception [3, 4], military security [5–7], etc. However, inadequate use of information, noise in the raw data, misalignment between individual sensors and different physical imaging principles are factors that have always led to limited detection performance. Hence, registration is an indispensable part of the applications of multimodal. Meanwhile, almost all of the existing multimodal techniques are designed for registered source inputs. As a result, these limitations require more complicated pre-registration multimodal imaging systems which always cost a lot, such as visible and infrared binocular systems [8–10]. To our best knowledge, there is still no registration-free multimodal detection method and benchmark aiming to implement high-quality detection using a simple multimodal imaging system so far.

Recently, a series of deep learning-based multimodal fusion and detection methods are proposed based on registered multimodal images captured by the pre-registration imaging system. DenseFuse [11] is an early work that plugs a densely connected network into the infrared and visible image fusion (IVIF) process as an extractor to fuse multimodal features; FusionGAN [12] presents an unsupervised and unified network for different types of multimodal fusion tasks; IFCNN [13] proposes a CNN-based fusion framework which inspired by the transform-domain conventional fusion algorithm; DDcGAN [14] is designed as a dual-discriminator GAN-style method for fusing infrared and visible images of different resolutions, and U2Fusion [15] solves different fusion tasks with a unified model and unified parameters including infrared and visible fusion. Then, the first combining implementation of IVIF and object detection TarDAL [10] is proposed after. Although these methods perform well on IVIF tasks, all of them require the raw multimodal data to be registered. Unfortunately, capture and registration have high requirements for multimodal imaging systems and image after-process.

To address the above concerns, we first constructed a simple visible-infrared multimodal imaging system containing two low-cost surveillance cameras and then captured an open-door multimodal smoke and wildfire detection benchmark named IA-VSW. Afterward, we also propose a CNN-Transformer hybrid learning framework to implement high-quality fusion detection using unregistered source images, namely the infrared-aware visible fusion detection network (IA-VFDnet). As shown in Fig. 1, the proposed IA-VSW contains two imaging settings: a large parallax setting and a small parallax setting. See Table 1 for IA-VSW detailed description.

**Table 1. IA-VSW Benchmark detailed description**

| Large parallax setting | | Small parallax setting | |
|---|---|---|---|
| Scenes: | 8 | Scenes: | 7 |
| Image Pairs: | 3167 | Image Pairs: | 3396 |
| Label Type: | VOC | Label Type: | VOC |
| Objects: | Smoke, Wildfire | Objects: | Smoke, Wildfire |
| Imaging Carma: HIKVISION EZVIZ C6Wi × 2 | | | |
| Image resolution: 1920×1080 | | | |

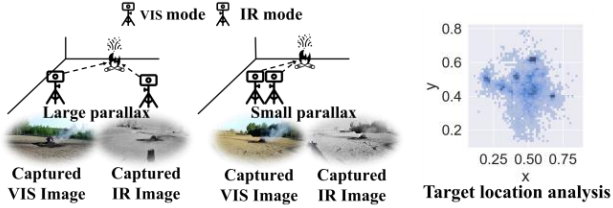

Fig.1. IA-VSW imaging systems & bounding box statistical analysis. The left one demonstrates the IA-VSW imaging system which contains two image settings: large parallax and small parallax; the right one shows the bounding box statistical distribution of the target position in the dataset.

Overall, the proposed IA-VFDnet (See Fig. 2) has a dual-branch encoder and a decoupled detection head. The RepVGG block [16] and Swin Transformer block [17] are used in our backbone. The RepVGG convolutional group is CNN-style in the sense that it adopts a plain topology and heavily uses 3 × 3 convolutions to model local features, and it is adopted into visible input processing as the main branch. The Swin-Transformer branch splits an infrared input into non-overlapping patches by patch splitting, and the transformer-style blocks adopt multi-head self-attention to model global features. For more details, we used 14, 24, and 1 RepVGG convolutional groups as RepVGG blocks, respectively. Meanwhile, we employed 2, 6, and 2 transformer groups with 8, 16, and 32 multi-head dimensions as Swim-Tr blocks, respectively. The embedding layer dimension in Swim-Tr was set to 128. Next, we used three coupled AKM and WDAM modules to integrate these shallow, middle, and deep features from different branches. Finally, we chose the YOLOX head [18] to implement detection.

Detailly, unregistered IA-VSW feature fusion always requires fusing features representing the same semantics from different sources. In essence, this is an optimal matching problem for bipartite graphs, and the most classical solution to this problem is the Kuhn-Munkres (KM) algorithm [19]. Therefore, we design AKM to improve the efficiency and quality of feature matching and to achieve an improved feature representation of the target information and we also design WDAF to denoise and fuse features.

As shown in Fig. 3 (a), the AKM module remodels the complementary feature $F_{li}^{\phi_y}$ referred on the fiducial feature $F_{li}^{\phi_x}$ to match features from different sources that represent homologous semantics.

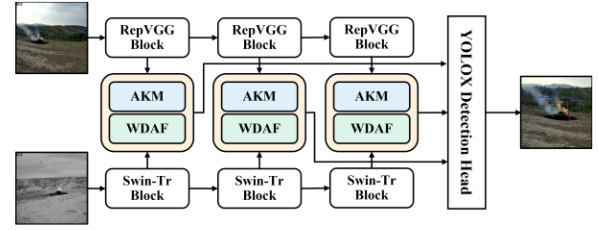

Fig.2. IA-VFDnet pipeline. Two input streams (visible and infrared) use RepVGG and Swin-transformer respectively as the backbone; the AKM and WDAF fuse multi-scale fiducial features extracted by RepVGG and complementary features extracted by Swin-transformer for object detection.

The features $F_{li}^{\phi_x} \in \mathbb{R}_x^{(c,w,h)}$ and $F_{li}^{\phi_y} \in \mathbb{R}_y^{(c,w,h)}$ are reshaped to $F_{li}^{\phi_x r} \in \mathbb{R}_x^{(c,w \times h)}$ and $F_{li}^{\phi_y r} \in \mathbb{R}_y^{(c,w \times h)}$, respectively. Then, for each fixed sub-feature $f_c^{\phi_x}$ on position c, we traversed the Chebyshev distances between sub-features $f_c^{\phi_x}$ and $f_{c1}^{\phi_y*}$ (the sub-feature of $F_{li}^{\phi_y*}$) to find out the homogeneous sub-feature pair $\{f_c^{\phi_x}, f_{c1}^{\phi_y*}\}$. Meanwhile, a learnable 1-D attention layer P consisting of a 1D convolutional layer is introduced to adaptively weigh the effective information of each complementary sub-feature, and the distance between the fiducial sub-feature and the complementary sub-feature is closed by the 1-D attention layer. The calculation of Chebyshev distance $\mathcal{D}(f_c^{\phi_x}, f_{c1}^{\phi_y*})$ is shown in Eq. (2). Finally, the mean value of sub-features $\{f_{c1}^{\phi_y*}, f_{c2}^{\phi_y*}, \dots, f_{cK}^{\phi_y*}\}$ can be calculated, which are chosen based-on the index of top K minimum $\mathcal{D}$. Thus, the matched feature $\vec{F}_{li}^{\phi_y}$ can be remodeled.

To fuse the fiducial feature $F_{li}^{\phi_x}$ with the remodeled complementary feature $\vec{F}_{li}^{\phi_y}$, We extended the wavelet transform to deep feature space in our proposed WDAF to achieve more fine-grained feature processing, as shown in Fig. 3 (b). These features are transformed into 3-D wavelet domain features by traversing each channel of the input 3D-feature using 2D-DWT. We also introduced multi-scale convolutions to fuse and denoise the wavelet domain concatenated feature. The parallel multi-scale convolutions can act as a denoiser during feature fusion. Next, a 1×1 convolution is used to integrate channels. Finally, the fused wavelet domain feature is split into 4 sub-band features for inverse DWT, and the fused feature $F_{li}^{\#}$ is obtained.

$$F_{li}^{\phi_y*} = P\left(F_{li}^{\phi_y r}\right) \quad (1)$$

$$\mathcal{D}\left(f_c^{\phi_x}, f_{c1}^{\phi_y*}\right) = \left| \| f_c^{\phi_x} \|_\infty - \| f_{c1}^{\phi_y*} \|_\infty \right| \quad (2)$$

where $F_{li}^{\phi_y*}$ represents the weighted complementary feature which contains sub-features $\{f_0^{\phi_y*}, f_1^{\phi_y*}, f_2^{\phi_y*}, \dots, f_c^{\phi_y*}\}$, $P$ denotes a learnable 1-D attention layer composed with 1-D convolution. Additionally, to better supervise the 1-D attention P and constrain the distance between the weighted complementary and the fiducial feature space distributions, the measurement loss $L_m$ is designed. Fig. 4 also shows the t-SNE visualization of $F_{li}^{\phi_x}$ and $F_{li}^{\phi_y}$. From a statistical view, $F_{li}^{\phi_x}$ and $F_{li}^{\phi_y}$ always have different distributions which has a negative effect on feature fusion. AKM aims to narrow the statistical gap between them. As previously stated, AKM remodels the feature by traversing and measuring the difference $\mathcal{D}$ of sub-features (the fiducial sub-features $f_i^{\phi_x}$ and the 1-D attention weighted the complementary sub-feature $\alpha_j \cdot f_j^{\phi_y}$). Therefore, the measurement loss $L_m$ is designed as Eq. (3).

$$L_m = \frac{1}{M \times N} \sum_{i=1}^{M} \sum_{j=1}^{N} \left| \| f_i^{\phi_x} \|_\infty - \| \alpha_j \cdot f_j^{\phi_y} \|_\infty \right| \quad (3)$$

where M and N represent the channel c of the fiducial feature $F_{li}^{\phi_x}$ and the complementary feature $F_{li}^{\phi_y}$.

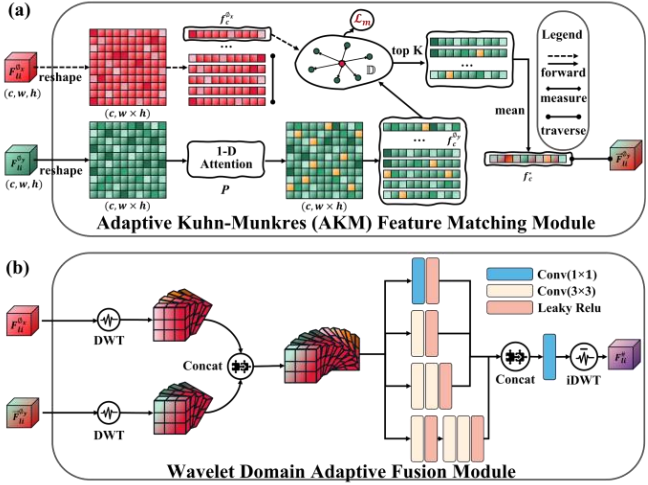

Fig.3. AKM and WDAF workflow. The complementary feature is remodeled according to the fiducial feature in (a). The fiducial feature and remodeling feature are transformed to the wavelet domain for denoising and fusing in (b).

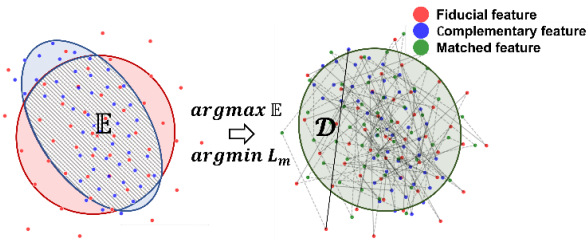

Fig.4. The t-SNE visualization of the distance between fiducial feature and complementary feature after remodeling.

To verify the performance of IA-VSW in multimodal fusion detection, we conducted experiments on two benchmarks: our proposed IA-VSW and M³FD [10]. IA-VSW focuses on smoke and wildfire detection with unregistered multimodal inputs, while M³FD focuses on general object detection with registered multimodal inputs. Additionally, our models were implemented using PyTorch and trained on 4 NVIDIA TESLA A40 GPUs for training. We used the Adam to optimize the model parameters; the momentum is 0.9 and the weight decay is 0.0005. The epoch was set to 100 with a batch size of 16, and the learning rate was 0.0001. We used the mosaic and mix-up to augment the training data

Firstly, six previously reported methods that combined fusion and detection networks were compared with our proposed IA-VFDnet on general object detection benchmark M³FD. The fusion methods contain DenseFuse, FusionGAN, IFCNN, DDcGAN, U2Fusion, and TarDAL. The YOLOX detection head was adopted as the detection network. For accuracy performance evaluation, we used the mean average precision (mAP@.50) metric [20], which measures detection performance across all categories. The separation proportion of the training and testing sets was 0.8:0.2 using shuffle sampling. As shown in Table 2, our proposed IA-VFDnet achieves the best performance on both six detection categories, and it performed 11.84% better than the second-place mAP@.50 metric.

Furthermore, we validate the performance of IA-VFDnet on smoke and wildfire detection benchmark IA-VSW. The separation proportion of the training and testing sets was 0.6:0.4 using shuffle sampling. The qualitative and quantitative results are shown in Fig. 5 and Table 3. It is observed that IA-VFDnet performed well on our proposed IA-VSW benchmark both in small and large parallax settings. It is noted that there is still no other method implementing smoke and wildfire detection under unregistered input conditions; thus, we only demonstrate the evaluation results of our proposed method.

To further demonstrate the superior performance of our proposed IA-VFDnet, we also compared the IA-VFDnet with SOTA methods under registered input conditions. The separation proportion of the training and testing sets was 0.8:0.2 using shuffle sampling, and the mAP@.50 metric was also used to evaluate the quantitative experiments. Fig. 6 shows some examples of comparisons.

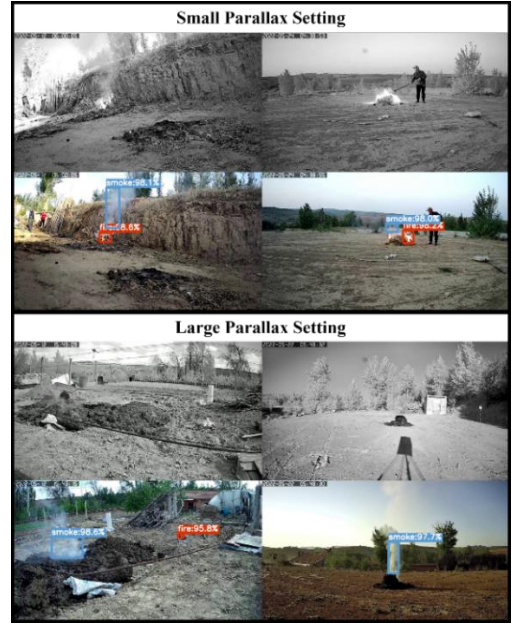

Fig.5. Results of IA-VFDnet on IA-VSW Benchmark under unregistered condition.

**Table 2. Comparisons with SOTA methods on M³FD under the registered condition**

| Method | M³FD Benchmark | | | | | | |
|---|---|---|---|---|---|---|---|
| | Person | Car | Bus | Motorcycle | Lamp | Truck | mAP@.50 |
| DenseFuse | 0.6558 | 0.7872 | 0.7520 | 0.6609 | 0.5724 | 0.6239 | 0.6753 |
| FusionGAN | 0.3121 | 0.6612 | 0.4752 | 0.3271 | 0.2468 | 0.3409 | 0.3938 |
| IFCNN | 0.7195 | 0.8639 | 0.7817 | 0.6718 | 0.6481 | 0.6848 | 0.7283 |
| DDcGAN | 0.4384 | 0.7737 | 0.6973 | 0.5399 | 0.3561 | 0.4994 | 0.5507 |
| U2Fusion | 0.6735 | 0.8720 | 0.8034 | 0.6929 | 0.7923 | 0.7235 | 0.7596 |
| TarDAL | 0.8503 | 0.8828 | 0.8351 | 0.7442 | 0.7905 | 0.7656 | 0.8114 |
| **Ours** | **0.9065** | **0.9091** | **0.9091** | **0.9026** | **0.9091** | **0.9091** | **0.9075** |

**Table 3. IA-VSW Performance Evaluation**

| Benchmark | IA-VFDnet performances on smoke and wildfire detection | | |
|---|---|---|---|
| | Smoke | Wildfire | mAP@.50 |
| IA-VSW (S) | 0.9088 | 0.8182 | 0.8635 |
| IA-VSW (L) | 0.9091 | 0.8177 | 0.8633 |

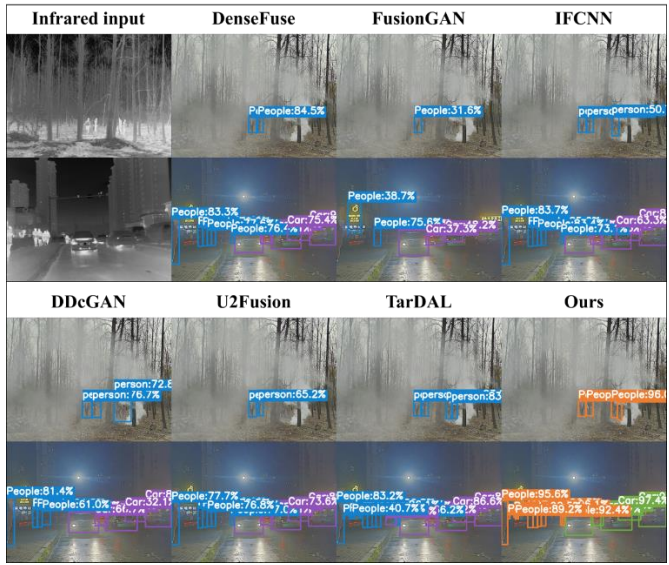

Fig.6. Results on M³FD Benchmark. Visual comparison of our IA-VFDnet and other SOTA methods under registered condition.

In this letter, an Infrared-Aware Visible Fusion Detection network (IA-VFDnet) is designed for low-cost, unregistered multimodal fusion detection. This network employs our designed Adaptive Kuhn-Munkres (AKM) and Wavelet Domain Adaptive Fusion (WDAF) modules to match and fuse deep features in the unregistered condition to prompt detection. Both qualitative and quantitative results prove the satisfactory performance of the proposed IA-VFDnet. To promote future research in infrared-aware visible fusion detection, this letter also opens the first infrared-aware visible smoke and wildfire detection benchmark, IA-VSW.

**Funding.** National Natural Science Foundation of China (61705092); State Key Laboratory for GeoMechanics and Deep Underground Engineering, China University of Mining & Technology (SKLGDUEK2118); Key Laboratory of Road Construction Technology and Equipment (Chang'an University) (MOE 300102251501).

**Disclosures.** The authors declare no conflicts of interest.

**Data availability.** The code and data underlying this letter is open available at https://github.com/yinghanguan/IA-VSW.